%% file: icme2025_template_anonymized.tex
\documentclass[conference]{IEEEtran}
\IEEEoverridecommandlockouts
\usepackage{cite}
\usepackage{amsmath,amssymb,amsfonts}
\usepackage{algorithmic}
\usepackage{graphicx}
\usepackage{textcomp}
\usepackage{booktabs}
\usepackage{multirow}
\usepackage{balance}
\usepackage[normalem]{ulem}
\useunder{\uline}{\ul}{}
\usepackage{caption}
\usepackage{subcaption}
\usepackage{pifont}
\newcommand{\xmark}{\ding{55}}%
\usepackage[table,xcdraw]{xcolor}
\usepackage[linesnumbered, ruled, vlined]{algorithm2e}
\usepackage{afterpage}
\usepackage[pagebackref,breaklinks,colorlinks]{hyperref}
\hypersetup{
    colorlinks=true,
    linkcolor=blue,
    urlcolor=blue
}

\def\BibTeX{{\rm B\kern-.05em{\sc i\kern-.025em b}\kern-.08em
    T\kern-.1667em\lower.7ex\hbox{E}\kern-.125emX}}

\begin{document}

\title{Interact with me: Joint Egocentric Forecasting of Intent to Interact, Attitude and Social Actions}


\author{
   Tongfei Bian$^1$, Yiming Ma$^2$, Mathieu Chollet$^1$, Victor Sanchez$^2$, and Tanaya Guha$^{1}$\\
   \textsuperscript{1}University of Glasgow, United Kingdom, \textsuperscript{2}University of Warwick, United Kingdom\\
}
\twocolumn[{
\renewcommand\twocolumn[1][]{#1}%
\maketitle
\vspace{-4mm}
\begin{center}
\captionsetup{type=figure}
\begin{subfigure}[t]{0.32\textwidth}
    \centering
    \includegraphics[width=\textwidth]{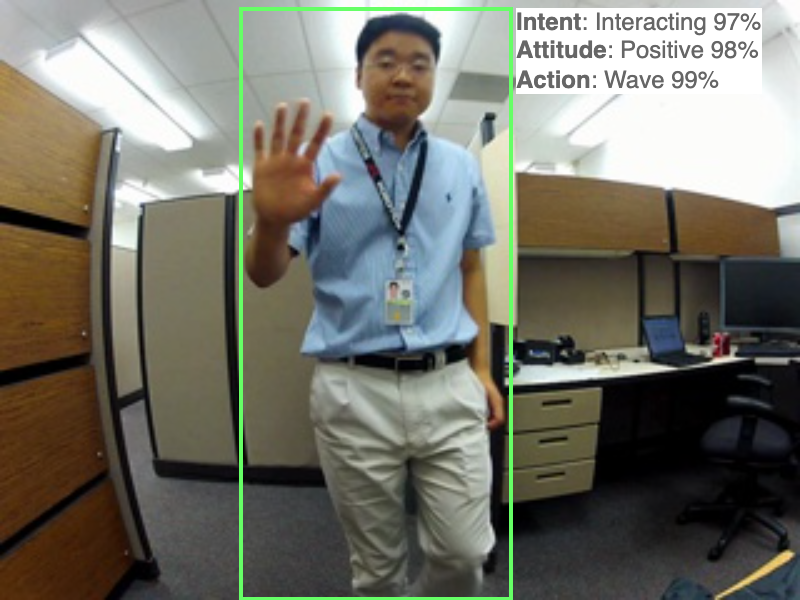}
    \caption{Positive interaction.}
    \label{fig:teaser_pos}
\end{subfigure}
\hfill
\begin{subfigure}[t]{0.32\textwidth}
    \centering
    \includegraphics[width=\textwidth]{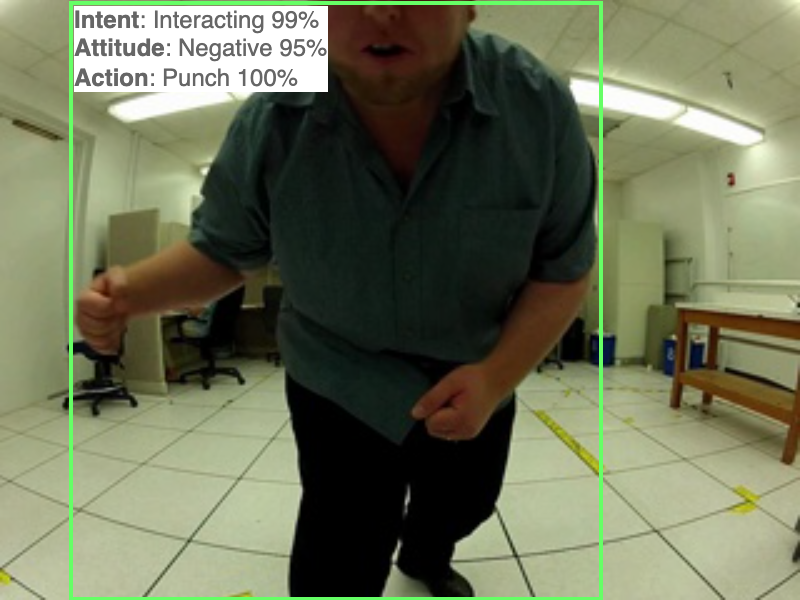}
    \caption{Negative interaction.}
    \label{fig:teaser_neg}
\end{subfigure}
\hfill
\begin{subfigure}[t]{0.32\textwidth}
    \centering
    \includegraphics[width=\textwidth]{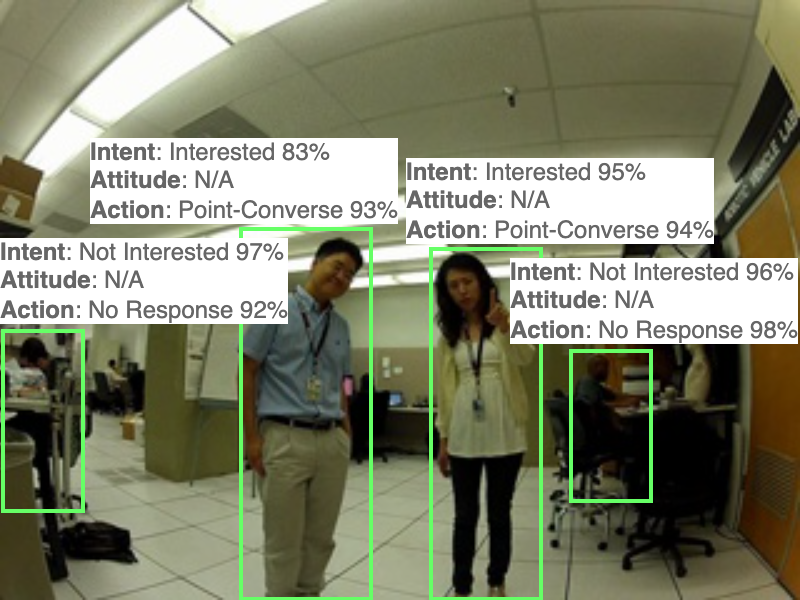}
    \caption{Interested or not to interact.}
    \label{fig:teaser_int}
\end{subfigure}
\caption{
To understand social interactions for human-agent interactions, we propose the task of jointly forecasting user's \textit{intent to interact}, their \textit{attitude}, and the \textit{action}. The intent-to-interact task forecasts whether a person in the field of view is \emph{interacting} (Fig.~\ref{fig:teaser_pos} and\ref{fig:teaser_neg}), \emph{interested} or \emph{not interested} (Fig.~\ref{fig:teaser_int}). Attitude towards agent is either positive (Fig.~\ref{fig:teaser_pos}) or negative (Fig.~\ref{fig:teaser_neg}). The action task anticipates which action the person is going to perform (all figures above).
}
\label{fig:teaser}
\end{center}
}]

\begin{abstract}
For efficient human-agent interaction, an agent should proactively recognize their target user and prepare for upcoming interactions. We formulate this challenging problem as a novel task of jointly forecasting a person's intent to interact with the agent, their attitude towards the agent and the action they will perform, from the agent's (egocentric) perspective. We propose \emph{SocialEgoNet} - a graph-based spatiotemporal framework that exploits task dependencies through a hierarchical multitask learning approach. SocialEgoNet uses whole-body skeletons (keypoints from face, hands and body) extracted from only 1 second of video input for high inference speed. For evaluation, we augment an existing egocentric human-agent interaction dataset with new class labels and bounding box annotations. Extensive experiments on this augmented dataset, named JPL-Social, demonstrate \emph{real-time} inference and superior performance (average accuracy across all tasks: 83.15\%) of our model outperforming several competitive baselines. The additional annotations and code are available at \href{https://github.com/biantongfei/SocialEgoNet}{\textbf{github.com/biantongfei/SocialEgoNet}}.
\end{abstract}


\input{1_introduction/main}
\input{2_literature/main}
\input{3_dataset/main}

%
\input{4_method/main}
%
\input{5_experiments/main}
\section{Conclusion}
\label{sec:conclusion}
We introduce a novel task for understanding social interactions from an egocentric perspective by jointly forecasting user's \emph{intent to interact}, their \emph{attitude} and \emph{actions}. To this end, we propose a new hierarchical multitask learning framework, \emph{SocialEgoNet}, that uses whole-body key points (face, hands, body) as input, and uses a spatiotemporal model for representation learning. This representation is input to a hierarchical multitask classifier that is designed to mimic human perception of interaction. We augmented the only existing egocentric interaction dataset with new labels and annotations. Our model achieved superior performance with high inference speed for all tasks. This makes our framework suitable for real-world deployment in human-robot interaction platforms. Future work will be directed towards incorporating multimodal data (gaze, audio, depth) and evaluating the system in-the-wild settings.

\balance
\bibliographystyle{IEEEbib}
\bibliography{egbib}
\vspace{12pt}
\end{document}

%% file: 1_introduction/main.tex
\section{Introduction}
\label{sec:intro}
Intelligent social agents (embodied or virtual) with a high degree of automation are now increasingly common in home environment and public spaces 
providing personal care assistance to navigation guidance \cite{intro_socialbot1_1}. In such situations, the social agents need to \emph{proactively} recognize their target user (often within a group of people) to prepare for upcoming interactions, rather than passively waiting for commands. This is critical to achieve smooth and effective human-agent interaction, because the agent's behavior in the early stage determines the success of such interactions \cite{intro_HRI}.

Consider a scenario where a human walks into a room full of people. By taking a quick glance, the person can easily identify with whom interaction is about to start, can even sense their attitude (positive or negative) or anticipate immediate action. For a robot/agent mimicking such behavior is a complex challenge. This requires analyzing the non-verbal behavioral cues of each person in the agent's field of view and forecast their \emph{intent to interact} with the agent. When a target user is identified, their \emph{attitude} and \emph{actions} also need to be forecast to enable timely and effective response. 
In particular, our framework identifies the following aspects of interaction at the early stage from an egocentric perspective:\\ \textit{(i) Who is going to interact with me? (ii) What is their attitude towards me? (iii) What action will they perform next?}
%
%

For a complex task like understanding interactions, the interplay among intent, attitude and actions need to be incorporated within the model itself - a challenge that our paper addresses.
We propose a novel egocentric framework called \textbf{SocialEgoNet} (see Fig. \ref{fig:model_structure} for an overview) to enable a human-like social understanding of interactions in visual scenes. The main idea is to extract high-level interaction-related information from low-level body, hand and facial key points (represented as pose graphs) for each person in the field of view and simultaneously perform multiple downstream forecasting tasks using a hierarchical structure \cite{podolak2008hierarchical}. 

Due to the unavailability of a suitable dataset, we augment the publicly available JPL-Interaction dataset \cite{JPL, jplextension} with person-level labelling of every subject's intent to interact and their attitude (see Section \ref{sec:data}). 
Our results show that SocialEgoNet can predict intentions with 88.10\% accuracy, attitudes with 91.11\% accuracy, and actions with 70.24\% accuracy. The average accuracy across all tasks is 83.15\%, 19.63\% higher than R3D-18 \cite{r3d} and 2.33\% higher than ST-GCN \cite{stgcn}. At the same time, SocialEgoNet has fewer parameters (66.24\% smaller than ST-GCN \cite{stgcn}) and faster inference speed (2.5x faster than ST-GCN \cite{stgcn}).
%
%
In summary, our contributions are:
\vspace{-1mm}
\begin{itemize}
    \item We introduce the novel task of jointly forecasting intent, attitude and actions from an egocentric view to understand social interactions. 
    \item We propose a new model called \textit{SocialEgoNet}. It achieves high accuracy across all tasks, outperforming competitive baselines despite being a smaller model, and achieving lower latency. We demonstrate that the hierarchical structure in SocialEgoNet can effectively learn complementary information between multiple downstream tasks.
     \item We augment the JPL dataset \cite{JPL,jplextension} (the only publicly available egocentric interaction dataset) with person-level annotations and new class labels. The new version will be made publicly available.
\end{itemize}
\input{figures/model_structure_2}

%% file: figures/model_structure_2.tex
\begin{figure*}[t]
    \centering
    \includegraphics[width=\textwidth]{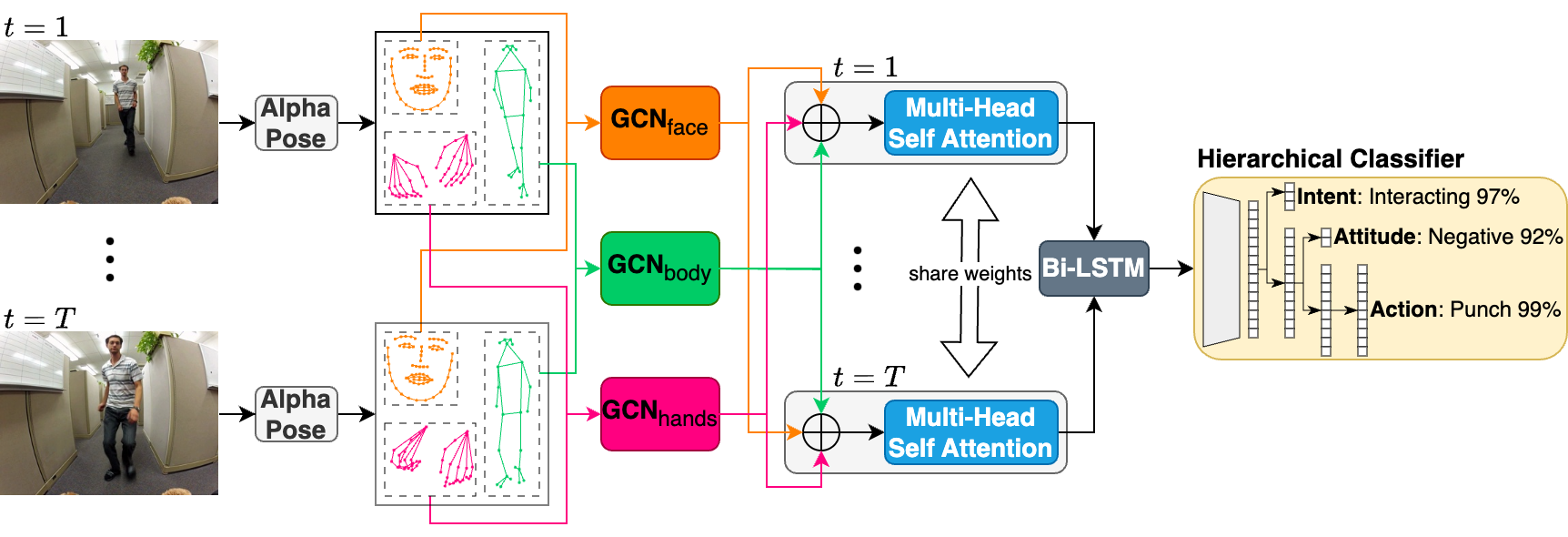}
    \caption{Overview of the proposed model, SocialEgoNet, for the joint forecasting task we introduce. 
    We first utilize AlphaPose to generate whole-body keypoints from the input video clip. The keypoints of the face, body, and hands are then fed into separate Graph Neural Networks (GCNs) to extract spatial representations, which are subsequently fused by concatenation and then passed through a multi-head self-attention module. The fused spatial representations from each frame are concatenated and the resulting sequence is fed into a bidirectional LSTM to model temporal relationships. Finally, the spatiotemporal feature output from the Bi-LSTM is passed to a hierarchical classifier to generate results for the three tasks.
    }
    \label{fig:model_structure}
    \vspace{-3mm}
\end{figure*}

%% file: 2_literature/main.tex
\vspace{-2mm}
\section{Related Work}
\label{sec:related_work}
\textbf{Intent and attitude prediction.} Modeling and predicting human intent based on visual input is critical for social agents to effectively serve or collaborate with humans \cite{human_intent}. In such tasks, nonverbal cues such as body pose, head movements, and facial expressions are crucial. Studies often focus on modeling the temporal evolution of these cues to understand the dynamic aspects of human intent \cite{relatedwork_nonverbal_cue2}.
%
Cues, such as gaze and motion have been used to analyse object manipulation intent \cite{relatedwork_intention1}.
%
Gaze and pose are also used to predict user engagement using a support vector classifier with a social robot \cite{brenner2021developing}. 
%
As for modeling the temporal evolution of the non-verbal cues, a Hidden Semi-Markov Model was used to reason about block placement intentions \cite{vinanzi2019mindreading}, while LSTM has been used to predict users intent to interact with a service robot \cite{abbate2024self}. In general, recurrent models are widely used and shown to capture the temporal patterns in non-verbal cues effectively.
%
Facial expressions and body gestures convey valuable information to perceive social intents and attitudes \cite{pereira2024systematic,fraser2022reducing}. CNNs have demonstrated strong performance on facial feature extraction \cite{souza2023grassmannian}. Also, CNNs can integrate facial and body pose features, with LSTM effectively capturing temporal dynamics for user attitude perception \cite{ilyas2021deep}.

\par\textbf{Action prediction.}
While action prediction is a well-studied area in computer vision, our focus is specifically on action prediction from an egocentric perspective. This task typically involves anticipating hand activities \cite{relatedwork_epickitchen1, grauman2022ego4d} and employs either LSTM or transformer-based approaches \cite{ajayi2024}.
LSTM-based methods often use rolling-unrolling techniques \cite{furnari2020rolling}, but they struggle to model long-horizon temporal dependencies. This limitation has been addressed by goal-based learning \cite{roy2022predicting} and memory-based modules \cite{wu2022memvit}.
Transformers are increasingly being used for action prediction \cite{grauman2022ego4d} due to their superior performance and the ease of integrating multiple modalities, especially text. The integration of large language models (LLMs) has been explored in several studies \cite{zhao2024antgpt, mittal2024can}, highlighting the potential for enhanced multimodal understanding in action anticipation. 

\textbf{Research gap.} Despite these advances, current approaches do not provide a comprehensive understanding of social interactions. So far, studies have considered individual aspects of the interaction in isolation. However, understanding interactions is a complex task and multiple aspects need to be considered simultaneously. Extracting high-level features that can inform multiple downstream tasks simultaneously also remains a significant challenge. We address these limitations by developing a holistic approach to understanding interaction in egocentric videos.

%% file: 3_dataset/main.tex
\vspace{-2mm}
\section{JPL-Social: Augmented JPL Dataset} 
\label{sec:data}


In this section, we first define the three joint tasks to forecast user's social intention from an egocentric perspective, and then introduce our \textbf{JPL-Social} Dataset, which is an augmented version of the JPL datasets \cite{JPL,jplextension}. We define the following forecasting tasks using only the first sec of video (see Fig.~\ref{fig:data}) to achieve a comprehensive understanding of egocentric interactions:
\begin{itemize}
    \item \textbf{Intent to interact}: Forecast whether or not any person in the view is \emph{interacting}, \emph{interested} or \emph{not interested} in interacting with the observer.
    \item \textbf{Attitude}: Forecast if the person's attitude towards the upcoming interaction is \emph{positive} or \emph{negative}.  
    \item \textbf{Action}: Forecast which type of action the person is going to perform. In this work, we focus on 10 action classes. 
\end{itemize}
%
\input{3_dataset/1_JPL}

\textbf{Why JPL?} This dataset is currently the only publicly available database focused on the recognition of egocentric interaction with robots. Although collected in a lab setting, it contains a variety of both positive and negative interactions with the opportunity to identify subjects interacting, but also those interested or not. However, the original dataset only contains labels at \emph{video-level} for the actions being performed by only the interacting subject. For the tasks defined above, the existing annotations are not adequate.
%
%
\input{tables/dataset_comparison}

\input{3_dataset/2_additional_ann}

%% file: 3_dataset/1_JPL.tex
%
\textbf{The JPL Dataset.} This dataset~\cite{JPL,jplextension} contains 200 egocentric videos of various actions performed by 16 subjects in front of an observer 
across 6 different settings with varying background and illumination. The videos are of resolution $320 \times 240$ pixels and recorded at 30 frames per second (FPS). The average length of a clip is $\sim$ 4.5 seconds.
Each video has one of the eight distinct interactions with the observer: `shaking hands', `hugging', `petting', `waving a hand', `having a conversation about the observer while occasionally pointing at it', `punching' `throwing objects' and `leaving'.
During interactions, the humanoid observer may also perform a large amount of ego-motion to imitate a social agent.

%% file: tables/dataset_comparison.tex
\begin{table}[t]
\begin{minipage}{\linewidth}
\caption{Comparison between the original JPL dataset and our JPL-Social which contains person-level annotations and new class labels. 
} 
\centering
\renewcommand*{\arraystretch}{1.2}
\resizebox{\columnwidth}{!}{%
\begin{tabular}{l|l|c|ccc}
\toprule
\multirow{2}{*}{\bf Dataset}    & \multirow{2}{*}{\bf Granularity} & \multirow{2}{*}{\bf Samples} & \multicolumn{3}{c}{\bf Number of Classes} \\ \cline{4-6} 
                            &                              &                                   & Intent     & Attitude     & Action    \\ \midrule
JPL \cite{JPL, jplextension}& video-level                  & 200                             & \xmark     & \xmark       & 8         \\
JPL-Social (ours)           & person-level                 & 290                          & 3          & 2            & 8+2       \\
  \bottomrule
\end{tabular}%
}
\label{tab:jpl_vs_jpl_social}
\end{minipage}
\begin{minipage}{\linewidth}
\centering
\vspace{5mm}
\includegraphics[width=0.6\linewidth]{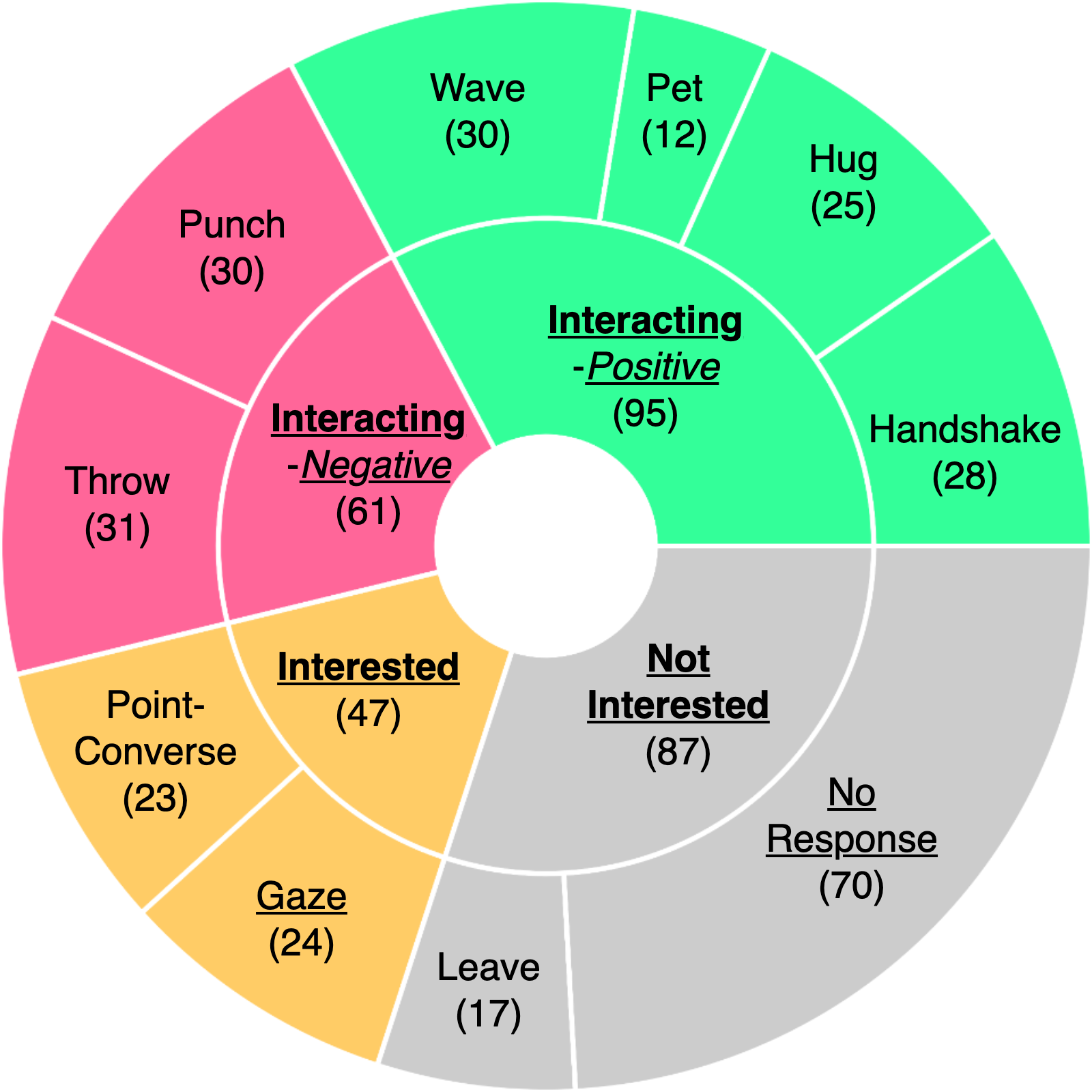}
\captionof{figure}{
\textbf{JPL-Social} class distribution for the three tasks—intent (in \textbf{bold}), attitude (in \textit{italic}), and actions (in regular). The number of samples in each class is indicated in parentheses, and newly introduced labels are \underline{underscored}.
}
\vspace{-6mm}
\label{fig:data}
\end{minipage}
\end{table}

%% file: 3_dataset/2_additional_ann.tex
\textbf{Additional Annotations.} The videos in the original dataset are labelled solely with the actions of the interacting subjects, while the actions of other individuals in the view are ignored. We augment the existing annotations with further fine-grained annotations for intent, attitude and additional actions for every person i.e., \emph{person-level} as opposed to video-level (see Table~\ref{tab:jpl_vs_jpl_social} and Fig.~\ref{fig:data} for more details). 

We use a pre-trained YOLOv3 \cite{redmon2018yolov3} to detect all visible subjects in every frame of a video. The tracks thus created (one for every person) are further annotated. The video-level labels provided originally are assigned to the interacting person directly. For the subjects not interacting, their tracks are annotated with \textbf{two new actions}: \emph{gaze} (i.e., looking at the observer) and \emph{no response} (i.e., focusing on their own business and ignoring the observer). Next, we annotate the \textbf{intent} and \textbf{attitude} of each subject. The intent of the interacting subject is labelled as `interacting'; for others we consider two labels: \emph{interested} and \emph{not interested}. The \emph{interested} label is assigned to subjects that are not yet interacting but are interested in interacting in the near future. A person is considered \emph{interested} when constantly looking at the observer (camera), otherwise, the person is labelled as \emph{not interested}. The \emph{not interested} label is assigned to subjects that completely ignore the observer. We create another set of labels that indicates the attitude of a subject towards the observer. This is done based on the actions they perform and applicable to only the 'interacting' subjects. Punching or Throwing an object to the observer are labelled as having a \emph{negative} attitude while the rest are considered as having a \emph{positive} attitude (see Fig.~\ref{fig:data}). Overall, we have 290 person-level tracks from all visible subjects in 200 video clips that are annotated  with three labels each: their intent to interact (3 classes), their attitude toward the observer (2 classes), and the (inter)action performed (10 classes).

\textbf{Data augmentation.} Given the small size of the dataset, we employ both frame-level and keypoint-level augmentation to augment the training set. At frame-level, each frame is randomly cropped three times at three different scales [0.95, 0.85, 0.75] before and after flipping. Next, we add Gaussian noise (0 mean, $\sigma$ standard deviation) to the keypoint locations as point-level augmentation, where $\sigma$ is proportional to the size of the annotated bounding box with a random scaling factor within [0.005, 0.01]. The augmentation steps increase the size of the training samples by two orders of magnitude.
%

%

%% file: 4_method/main.tex
\section{Approach}
\label{sec:methodology}
%
%
In this section, we describe our hierarchical multitask forecasting framework, named \textbf{SocialEgoNet} (see Fig.~\ref{fig:model_structure}),
which is capable of handling the three proposed tasks simultaneously. SocialEgoNet comprises (i) a pose estimation module, (ii) a spatiotemporal representation learning module, and (iii) a hierarchical multitask classifier. SocialEgoNet takes a video clip as input, and extracts whole-body (face, hand and body) key points for each person in the view using the pose estimation module. The key points are then fed to the spatiotemporal representation learning module, which analyzes the spatial information individually, fuses them, and finally learns the temporal dependency. The output embeddings are passed to the hierarchical multitask classifier to generate predictions for \emph{future} intent, attitude and actions.
%
%
\input{4_method/1_pose_estimation}
\input{4_method/2_spatiotemporal_model}
\input{algorithms/gcn_lstm}
\input{figures/mlp}
\input{tables/sota}
\input{4_method/3_hierarchical_classifier}
%
%

%% file: 4_method/1_pose_estimation.tex
\subsection{Whole-Body Pose Estimation Module}

We establish our framework on whole-body keypoints rather than RGB frames to (i) focus more on the user's body language, which is more important in social interactions than the surrounding environment; (ii) reduce bias such as skin colour, clothing texture and lighting; and (iii) require less computational resources and can be deployed on resource-constrained social agent platforms. 
%

We utilize AlphaPose \cite{alphapose} pre-trained on 
COCO2017~\cite{coco2017}  to estimate coordinates and confidence scores of whole-body keypoints (68 points for the face, 23 points for the body, and 42 points for the hands) for all visible humans in a video frame. Then, we construct adjacency matrices in COCO format \cite{coco2017} to connect the keypoints to form a pose graph.

%% file: 4_method/2_spatiotemporal_model.tex
\subsection{Spatiotemporal Representation Learning}
\label{sec:spatiotemporal}
For a sequence of whole-body pose skeleton graphs, we analysed the spatial representations of the \emph{body}, \emph{face} and \emph{hands} with three separate three-layer GCNs \cite{methodology_gcn}. Subsequently, the representations from the different body parts were combined and passed into a Multi-Head Self-Attention (MSA) \cite{methodology_transformer} to learn the relative importance of keypoints. After obtaining the spatial representations at each time step, we fed them into a three-layer Bi-LSTM \cite{methodology_blstm} to model temporal dependencies. See Algorithm~\ref{alg:gcn_lstm} for a summary.

%% file: algorithms/gcn_lstm.tex
\begin{algorithm}[t]
\SetKwData{Left}{left}
\SetKwData{This}{this}
\SetKwData{Up}{up}
\SetKwFunction{Union}{Union}
\SetKwFunction{FindCompress}{FindCompress}
\SetKwInOut{Input}{Input}\SetKwInOut{Output}{Output}
\SetKwComment{comment}{\#}{}

\Input{Keypoint coordinates $\{ \boldsymbol{x}_i^{(t)} \}_{t = 1}^T$, adjacency matrices $\boldsymbol{A}_i$ and confidences scores $\{ c_i^{(t)} \}_{t = 1}^T$, for $i = \text{b}, \text{f},\, \text{h}$, indicating the body part.}
\Output{The spatiotemporal representation vector $\boldsymbol{z}$.}
\BlankLine

\For{$t = 1 ,\, \cdots ,\, T$}{
    \For{$i = \text{b}, \, \text{f}, \, \text{h}$}{
        Build the graph $\boldsymbol{g}_i^{(t)}$ using $\mathcal{G}_i^{ (t) } = \mathcal{G}(\boldsymbol{x}_i^{ (t) },\, c_i^{ (t) }, \boldsymbol{A}_i)$.
        \\
        Extract spatial representations by 
        $\boldsymbol{z}_i^{(t)} = \texttt{GCN}_i (\mathcal{G}_i^{(t)})$.
    }
    Concatenate $\boldsymbol{z}_\text{b}^{(t)}$, $\boldsymbol{z}_\text{f}^{(t)}$, $\boldsymbol{z}_\text{h}^{(t)}$ using 
    $\boldsymbol{z}^{(t)} = \texttt{Concat} (\boldsymbol{z}_\text{b}^{(t)} ,\, \boldsymbol{z}_\text{f}^{(t)} ,\, \boldsymbol{z}_\text{h}^{(t)} ) \in \mathbb{R}^{n \times 16}$\\
    Pass the combined feature $\boldsymbol{z}^{(t)}$ to \texttt{MSA} 
    \\
    Flatten the result to obtain $\bar{\boldsymbol{z}}^{(t)}$.
}
Group all $\bar{\boldsymbol{z}}^{(t)}$ into a sequence $\{\bar{\boldsymbol{z}}^{(t)} \}_{t = 1}^T$.\\
Feed the sequence into \texttt{Bi-LSTM} to get $\{\tilde{\boldsymbol{z}}^{(t)}\}_{t = 1}^T$.\\
Output $\tilde{\boldsymbol{z}}^{(T)}$ as spatiotemporal feature $\boldsymbol{z}$.

\caption{Pseudocode for extracting spatiotemporal representations in SocialEgoNet.
}
\label{alg:gcn_lstm}
\end{algorithm}
\vspace{-2mm}

%% file: figures/mlp.tex
\begin{figure}[t]
    \centering
    \includegraphics[width=\linewidth]{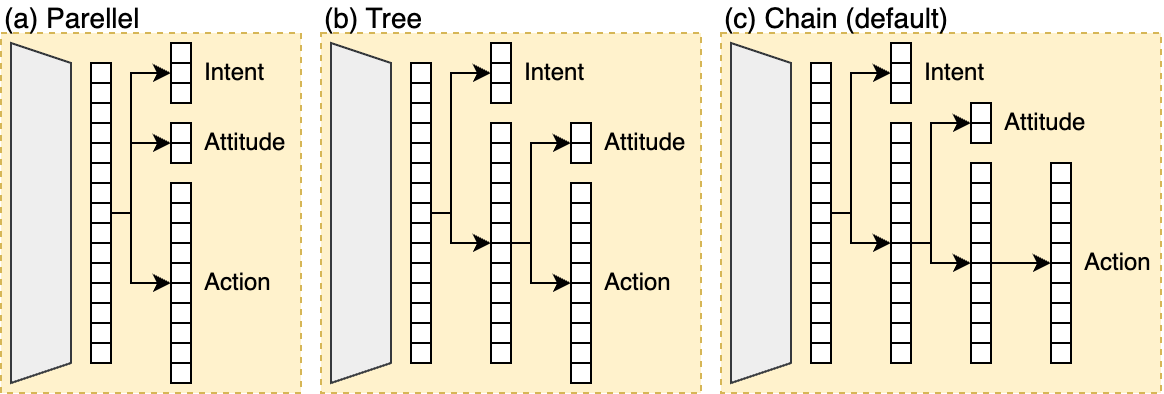}
    \caption{The proposed designs of our hierarchical classifiers. 
    }
    \label{fig:mlp}
    \vspace{-5mm}
\end{figure}

%% file: tables/sota.tex
\begin{table*}[t]
\centering
\caption{Comparison of SocialEgoNet with competitive spatiotemporal models on the JPL-Social dataset by observing the first second of interaction (30 frames). The best results are in \textbf{bold}, while the second-best results are \underline{underlined}. 
``$\Delta$'' refers to the time to extract the whole-body pose keypoints. AlphaPose \cite{alphapose} takes 1.27 ms to extract whole-body pose features from an annotated frame. All the experiments run on a RTX 4070.}
\renewcommand*{\arraystretch}{1.2}
\small
\resizebox{\textwidth}{!}{%
\begin{tabular}{l|cc|cccccc|cc}
\toprule
\multirow{2}{*}{\textbf{Method}} &
  \textbf{Params} &
  \textbf{Latency} &
  \multicolumn{2}{c}{\textbf{Intent}} &
  \multicolumn{2}{c}{\textbf{Attitude}} &
  \multicolumn{2}{c|}{\textbf{Action}} &
  \multicolumn{2}{c}{\textbf{Average}} \\
 &
  (M) $\downarrow$ &
  (ms) $\downarrow$ &
  \textbf{F1} (\%) &
  \textbf{Acc} (\%) &
  \textbf{F1} (\%) &
  \textbf{Acc} (\%) &
  \textbf{F1} (\%) &
  \textbf{Acc} (\%) &
  \textbf{F1} (\%) &
  \textbf{Acc} (\%) \\ \midrule
R3D-18 \cite{r3d} &
  33.17 &
  8.67 &
  75.14 &
  75.61 &
  70.02 &
  67.39 &
  46.41 &
  47.56 &
  63.86 &
  63.52 \\
ST-GCN \cite{stgcn} &
  9.42 &
  {\ul $\Delta$ + 1.40} &
  87.30 &
  86.90 &
  87.84 &
  88.89 &
  65.19 &
  66.67 &
  80.11 &
  80.82 \\
DG-STGCN \cite{dgstgcn} &
  {\ul 8.78} &
  $\Delta$ + 3.47 &
  83.92 &
  83.33 &
  84.34 &
  89.36 &
  67.18 &
  67.86 &
  78.48 &
  80.18 \\
MS-G3D \cite{msg3d} &
  12.82 &
  $\Delta$ + 4.74 &
  \textbf{90.02} &
  \textbf{89.29} &
  \textbf{90.11} &
  \textbf{93.33} &
  \textbf{73.29} &
  \textbf{72.62} &
  \textbf{84.47} &
  \textbf{85.08} \\
\textbf{SocialEgoNet (ours)} &
  \textbf{3.18} &
  \textbf{$\Delta$ + 0.56} &
  {\ul 88.43} &
  {\ul 88.10} &
  {\ul 88.99} &
  {\ul 91.11} &
  {\ul 69.57} &
  {\ul 70.24} &
  {\ul 82.33} &
  {\ul 83.15} \\ \bottomrule
\end{tabular}%
}
\label{tab:spatiotemporal_models}
\vspace{-2mm}
\end{table*}

%% file: 4_method/3_hierarchical_classifier.tex
\subsection{Hierarchical Classifier}
To perform the three tasks simultaneously, we feed the spatiotemporal embedding $\boldsymbol{z}$ to a dense neural network (DNN) with three heads. Our hierarchical classifier is designed to mimic the steps of human perception while anticipating an interaction. We consider three different hierarchical designs (see Fig.~\ref{fig:mlp}) for this module:
\begin{itemize}
    \item \textbf{Parallel}: Treats the tasks independently, so the three prediction heads are treated equally and operate in parallel.
    \item \textbf{Tree}: Considers that social interest can be regarded as the superclass of attitudes and actions. 
    Therefore, the prediction for social interest is generated first, followed by the simultaneous predictions for attitudes and actions. 
    \item \textbf{Chain} (default): Build on the tree structure, this design further assumes that attitude can be regarded as the parent class of some actions. Thus, it generates predictions sequentially: first for interests, then for attitudes, and finally for actions. This structure imitates how humans sense others in the real world. 
\end{itemize}



%% file: 5_experiments/main.tex
%
\section{Experiments}
\label{sec:experiments}
\input{5_experiments/1_sota}
\input{5_experiments/2_contribution_of_body_parts}
\input{tables/body_parts}
\input{5_experiments/3_classifiers}
\input{tables/hierarchical_classifiers}
%
\input{figures/diff_window_sizes}
\input{5_experiments/5_observation_window}

%% file: 5_experiments/1_sota.tex
\subsection{Comparisons with State-of-the-Art}
%
%
We compare SocialEgoNet with the following state-of-the-art models (results in Table \ref{tab:spatiotemporal_models}): 
\\
$\bullet$ Graph-based models i.e., ST-GCN \cite{stgcn}, MS-G3D Net \cite{msg3d} and DG-STGCN \cite{dgstgcn}. The inputs to these methods are the face, body and hand keypoints - the same as for SocialEgoNet. 
We assign a network to each body part separately to learn spatiotemporal features and feed the spliced features into a chain hierarchical classifier.

\noindent $\bullet$ 3D-CNN-based R3D-18 \cite{r3d}: We first extract the track of each person in the video using our annotated boxes (zero padding is utilized to align the spatial size). The tracks are fed to the R3D backbone to extract spatiotemporal features, which is passed to the chain hierarchical classifier. 

Model sizes and inference speeds are critical for social agents; therefore we also report each model's number of parameters 
and latency (the average time required to infer from 30 frames of people's annotated boxes)
. Table \ref{tab:spatiotemporal_models} demonstrates that
(1) Graph-based approaches, including SocialEgoNet, have smaller model sizes and faster inference speed while consistently outperforming the 3D-CNN model R3D across the three tasks, demonstrating their superior data and computational efficiency. (2) MS-G3D achieves the best performance in all tasks, but our model, SocialEgoNet, ranks second with marginal gaps in Intent and Attitude prediction tasks. This is encouraging as SocialEgoNet is designed with a smaller model size (75.20\% smaller than MS-G3D) and a lower latency (8.5x faster than MS-G3D). (3) Compared to Intent and Attitude, Action seems more challenging due, as it requires recognizing subtle differences in similar actions. For example, hugging and petting may share the same initial hand movement. 

%% file: 5_experiments/2_contribution_of_body_parts.tex
\subsection{Contribution of Body Parts}
Next we analyse the contribution of each body part towards interaction perception by training SocialEgoNet with different combinations of keypoints as input. Table \ref{tab:body_parts} shows that, when comparing individual body parts, face carries the most information for forecasting intent and attitude, while body and hand can better forecast actions. When face is combined with body or hands, SocialEgoNet's performance on action prediction is improved from 49.41\% F1 score to at most 64.89\%. The best performance for all three tasks is achieved when all the three body parts are used as inputs. As more body parts are analysed, performance across tasks improves which means that each body part provides irreplaceable and complementary information.

%% file: tables/body_parts.tex
\begin{table}[t]
\centering
\caption{Significance of different body parts: F1 score (in $\%$) is used as an evaluation metric and the best results are in \textbf{bold} font. In most cases, including information from multiple body parts leads to improved performance as \colorbox{blue!10}{highlighted}.}
\resizebox{0.9\columnwidth}{!}{%
\begin{tabular}{l|ccc|c}
\toprule
\textbf{Body Parts}          & \textbf{Intent} & \textbf{Attitude} & \textbf{Action} & \textbf{Average} \\ \midrule
Body & 83.41 & 79.01 & 63.57 & 75.33 \\
Face & 86.23 & 83.38 & 49.41 & 73.01 \\
Hands & 68.06 & 76.55 & 62.05 & 68.89 \\
\midrule
Body + Face & 86.12 & \cellcolor{blue!10}83.99 & \cellcolor{blue!10}64.89 & \cellcolor{blue!10}78.33 \\
Body + Hands   & \cellcolor{blue!10}84.29 & \cellcolor{blue!10}80.79 & \cellcolor{blue!10}67.16 & \cellcolor{blue!10}77.41 \\
Face + Hands   & 85.83 & \cellcolor{blue!10}85.06 & \cellcolor{blue!10}63.17 & \cellcolor{blue!10}78.02 \\
\midrule
\textbf{All} & \cellcolor{blue!10}\textbf{88.43}  & \cellcolor{blue!10}\textbf{88.99}    & \cellcolor{blue!10}\textbf{69.57}  & \cellcolor{blue!10}\textbf{82.33}   \\ 
\bottomrule
\end{tabular}%
}
\label{tab:body_parts}
\vspace{-2mm}
\end{table}

%% file: 5_experiments/3_classifiers.tex
\vspace{-1mm}
\subsection{Hierarchy Design}
%
In this section, we study the effects of different designs of the hierarchical classifiers (see Fig.~\ref{fig:mlp}). Recall that this hierarchical design attempts to mimic how humans perceive and anticipate interactions. Table \ref{tab:hierarchical_classifiers} shows 
that SocialEgoNet performs the best with Chain structure for all tasks, and therefore is set as the default in the SocialEgoNet. We also performed classification focusing on a single task separately to compare the gains of hierarchical classifiers. We note performance gain in every case, reinforcing the need for a hierarchical classification design and proving that it can learn the interaction between different tasks.

%% file: tables/hierarchical_classifiers.tex
\begin{table}[t]
\centering
\caption{Comparison of different hierarchical classifiers in SocialEgoNet. The F1 score (in percent) is used as the evaluation metric, and the best results are shown in \textbf{bold}.}
\resizebox{0.85\columnwidth}{!}{%
\begin{tabular}{l|ccc|c}
\toprule
\textbf{Classifier}          & \textbf{Intent} & \textbf{Attitude} & \textbf{Action} & \textbf{Average} \\ \midrule
Single task & 86.17 & 86.15 & 64.48 & 78.93 \\
\midrule
Parallel & 86.45 & 87.72 & 66.21 & 80.13 \\
Tree     & 86.10 & 87.35 & 67.28 & 80.24 \\
\textbf{Chain} & \textbf{88.43}  & \textbf{88.99}    & \textbf{69.57}  & \textbf{82.33}   \\ 

\bottomrule
\end{tabular}%
}
\label{tab:hierarchical_classifiers}
\end{table}

%% file: figures/diff_window_sizes.tex
\begin{figure}[tb]
    \centering   \includegraphics[width=0.74\linewidth]{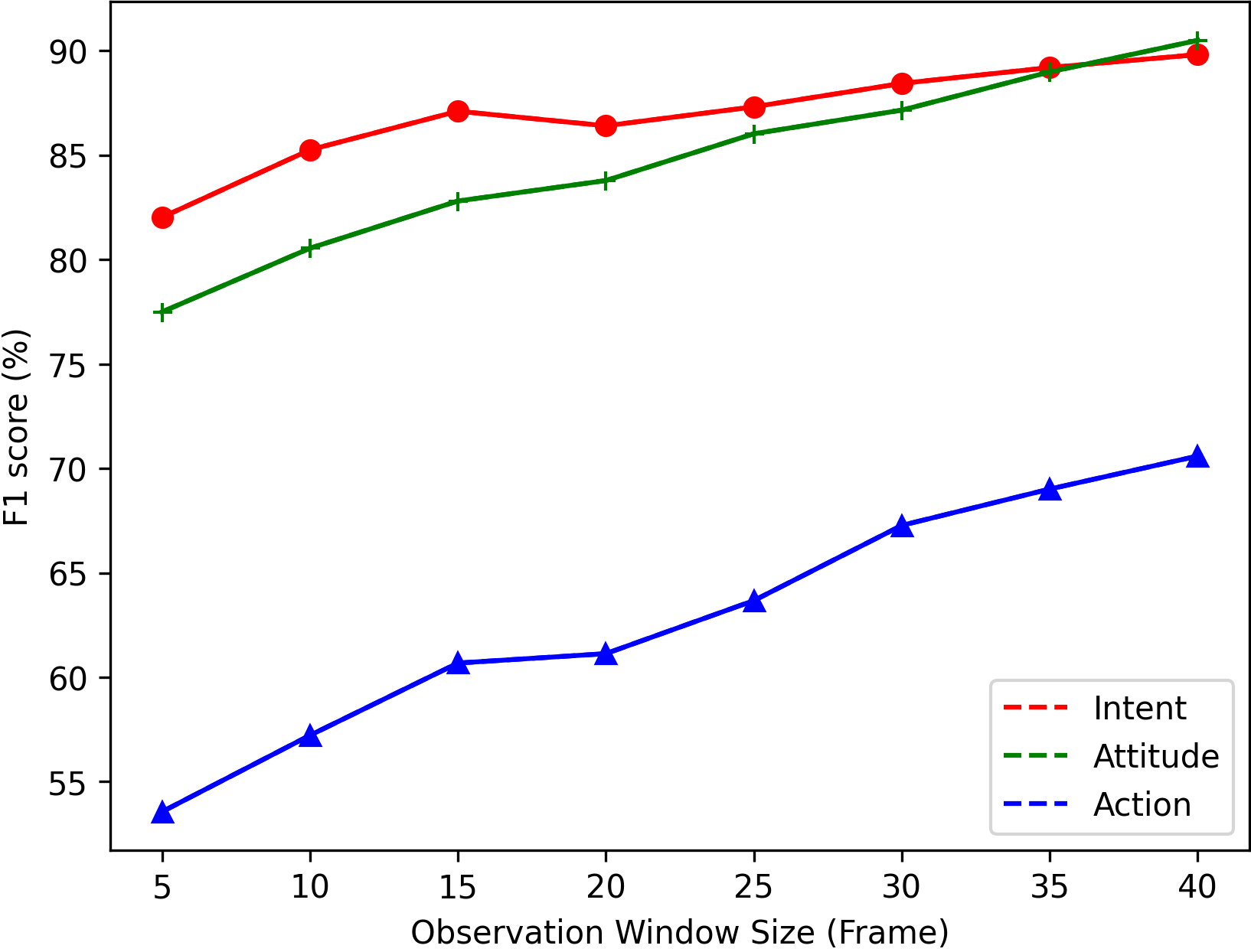}
    \caption{Effect of the observation window size on SocialEgoNet's performance. }
    \label{fig:observe_window}
\vspace{-5mm}
\end{figure}

%% file: 5_experiments/5_observation_window.tex
\vspace{-1mm}
\subsection{Impact of Observation Window Size}
\vspace{-1mm}
We also investigate how the observation window size (i.e., how many frames are used as input) impacts the performance. Fig.~\ref{fig:observe_window} shows that in general performance increases with window size, since more information is available. However, the effects of this parameter on different tasks are different. For intent, larger window size improves the performance, while for attitude and action tasks the effect is more pronounced. This reflects the varying levels of challenge the tasks pose, i.e., identifying people intend to interact takes a shorter observation time, while accurately forecasting their attitude and future actions takes longer. 